%% file: pruning_paper.tex
\documentclass{article}
\newcommand\arxiv{1}
\ifnum \arxiv>0
    \usepackage[preprint]{neurips_2020}
\else
    \usepackage{neurips_2020}
\fi
\usepackage[utf8]{inputenc} 
\usepackage[T1]{fontenc}    
\usepackage{hyperref}       
\usepackage{url}            
\usepackage{nicefrac}       
\usepackage{wrapfig}
\usepackage[x11names,dvipsnames,table]{xcolor}
\usepackage{microtype}
\usepackage{graphicx}
\usepackage{subcaption} 
\usepackage{amssymb}
\usepackage{pifont}
\newcommand{\cmark}{\ding{51}}%
\newcommand{\xmark}{\ding{55}}%
\usepackage{multirow}
\usepackage{makecell} 
\usepackage{tikz}
\usetikzlibrary{shapes,arrows}
\usepackage{pgfplots}
\usepackage{adjustbox}
\usepackage{amsmath}
\usepackage{amsfonts}
\usepackage{amssymb}
\usepackage{algorithmic}
\usepackage{textcomp}
\usepackage{booktabs} 
\usepackage{hyperref}
\usepackage{colortbl}
\usepackage[dvipsnames]{xcolor}
\usepackage{float}
\restylefloat{table}
\RequirePackage{algorithm}
\RequirePackage{algorithmic}
\include{prel}

\everymath{\displaystyle\everymath{}}
\pgfplotsset{compat=1.14}
\usepackage{arydshln}

\title{Knapsack Pruning with Inner Distillation}

\author{%
  Yonathan Aflalo\thanks{Corresponding author} \\
  Alibaba Damo Academy\\
  Tel Aviv, Israel \\
  \texttt{Johnaflalo@gmail.com} \\
  \And
  Asaf Noy \\
  Alibaba Damo Academy\\
  Tel Aviv, Israel \\
  \And
  Ming Lin \\
  Alibaba Damo Academy\\
  Seattle, USA \\
  \And
  Itamar Friedman \\
  Alibaba Damo Academy\\
  Tel Aviv, Israel \\
  \And
  Lihi Zelnik \\
  Alibaba Damo Academy\\
  Tel Aviv, Israel \\
}

\begin{document}

\maketitle

\begin{abstract}
Neural network pruning reduces the computational cost of an over-parameterized network to improve its efficiency. Popular methods vary from $\ell_1$-norm sparsification to Neural Architecture Search (NAS).  In this work, we propose a novel pruning method that optimizes the final accuracy of the pruned network and distills knowledge from the over-parameterized parent network's inner layers. To enable this approach, we formulate the network pruning as a Knapsack Problem which optimizes the trade-off between the importance of neurons and their associated computational cost. Then we prune the network channels while maintaining the high-level structure of the network. The pruned network is fine-tuned under the supervision of the parent network using its inner network knowledge, a technique we refer to as the {\it Inner Knowledge Distillation}. Our method leads to state-of-the-art pruning results on ImageNet, CIFAR-10 and CIFAR-100 using ResNet backbones. 
To prune complex network structures such as convolutions with skip-links and depth-wise convolutions, we propose a block grouping approach to cope with these structures.
Through this we produce compact architectures with the same FLOPs as EfficientNet-B0 and MobileNetV3 but with higher accuracy, by $1\%$ and $0.3\%$ respectively on ImageNet, and faster runtime on GPU.

\end{abstract}
\section{Introduction}

Deep and wide networks such as VGG~\cite{VGG}, ResNet~\cite{ResNet} and EfficientNet~\cite{EfficientNet}  achieve high classification accuracy on challenging benchmarks such as ImageNet~\cite{imagenet_cvpr09}. 
While these architectures perform well, in many scenarios it is desired to reduce their computational cost and model size.
One approach to achieve this goal is via network pruning which has been a topic of research for decades~\cite{braindamage}.
Network pruning is a way to identify and remove the insignificant parameters of a network.
These are the ones with little effect on the accuracy.

Previous pruning methods show promising results. However, they suffer from two key shortcomings.
The first is the relying on a coarse approximation of the contribution of each weight on the final accuracy.
For example, NetAdapt~\cite{DBLP:conf/eccv/YangHCZGSSA18} measures the post-factum empirical influence of several pruning options in order to choose the best one.
The second is not leveraging the expressive power of the parent network. 
Knowledge Distillation (KD)~\cite{KnowledgeDistillation} from the unpruned network could improve performance as shown by~\cite{TAS} who used KD on the network outputs to fine-tune the child network. Their approach, however, does not leverage the fact to the full extent that the inner structures of the unpruned and pruned networks are highly isomorphic.

In this paper we present a pruning approach that optimizes explicitly on the trade-off between accuracy and computational cost.
Our first key idea is to formulate the pruning as a Knapsack Problem which enables the trade-off optimization.
The second key idea is to introduce an {\it Inner Knowledge Distillation} (IKD) mechanism between the inner layers of the pruned and unpruned network. The IKD guides the child network to reproduce the inner layer's mapping patterns of the unpruned parent network as much as possible, leading to higher accuracy after fine-tuning.

The integration of the above two key ideas allows us to develop a novel method with strong empirical performance.
Our method is a one-shot method, it is fast and does not require iterative re-training during pruning. 
The Knapsack formulation we suggest enables the pruning of non-sequential convolutions such as skip-connections and Squeeze-and-Excitation modules which are common in modern architectures, for example, ResNet and EfficientNet~\cite{EfficientNet}. We show that our method leads to state-of-the-art results on ImageNet, CIFAR-$10$ and CIFAR-$100$ when using ResNets and EfficientNets as backbones.  

The structure of the paper is as follows: In Section \ref{sec:previous}, we briefly review previous works on pruning and knowledge distillation. In Section \ref{sec:methodology_seq}, we describe the technical aspects of our method to prune sequential convolutions or convolutions that are not connected to a skip-connection. In Section \ref{sec:gmcc}, we extend our method to more complicated architectures, that include skip-connections, dilated convolutions or Squeeze-and-Excitation modules which enforce constraints on the convolutional channels to be pruned. In Section \ref{sec:IKD}, we describe our fine-tuning method with IKD. Finally, in Section \ref{sec:exp}, we present the results of our pruning method on different benchmarks and backbones.

\section{Related Works}
\label{sec:previous}

In this section, we briefly review previous works on pruning and knowledge distillation that closely relate to our work.
\subparagraph{Network Pruning}
Network pruning dates back to~\cite{braindamage} where the importance of a neuron is estimated by the diagonal elements of the Hessian matrix of the network's loss function. For modern neural networks estimating the Hessian matrix is prohibitive due to the high dimensionality. Therefore, inspired by the success of compressed sensing techniques ~\cite{donoho2006compressed}, many $\ell_1$-norm sparsification methods and sparse proximal projection methods have been introduced to prune over-parameterized networks~\cite{conf/cvpr/LiuWFTP15,ding2019global,liu2017learning}. 
These methods require iterative pruning during training which makes them inapplicable to pre-trained networks. 

Methods that perform post-training pruning over pre-trained neural networks are under active research recently~\cite{NIPS2015_5784,PFEC,DBLP:conf/eccv/YangHCZGSSA18,NISP,DBLP:journals/corr/HuPTT16}. Their key idea is to estimate the importance of a neuron via some heuristics. A comprehensive comparison of pruning heuristics is presented in ~\cite{DBLP:conf/iclr/MolchanovTKAK17}, including Minimum $\ell_2$ Weight, Activation, Mutual Information, Taylor Expansion, and Average Percentage of Zeros. They show that the best criterion is the Taylor Expansion which approximates the change in the loss function induced by the pruning. 

More recently, ~\cite{molchanov2019taylor} demonstrated the high correlation between the importance approximation to a reliable estimate of the true importance of the neurons. However, their decision of removing $N$ neurons with the smallest importance scores is rather heuristic and does not account for the induced change of FLOPs.
\subparagraph{Knowledge Distillation}
Knowledge distillation refers to training a student network using a teacher network by distilling information from the teacher to the student. \cite{KnowledgeDistillation} uses a penalty term consisting of the cross entropy between the output logits of the teacher and that of the student in the loss function.
A few methods use knowledge distillation inside the network. For example, ~\cite{li2019blockwisely, progressiveblock} consider the $\ell_2$ distance between the teacher and the student feature maps as part of the loss.
When the dimensions of the feature maps of the two networks differ, a popular method is to penalize the distance between the embeddings of the features maps in a subspace of lower dimension. For instance, ~\cite{moonshine} computes the $\ell_2$ distance between the squared sum of the teacher and the student feature maps while ~\cite{Tung2019SimilarityPreservingKD} penalizes the distance between the activation correlation matrices. A distillation at the level of the feature maps has been already studied by previous works such as \cite{fitnets, Heo_2019_ICCV}, but the internal feature maps on which the distillation is performed are chosen arbitrarily.
\subparagraph{Knapsack Problem} The knapsack problem is extensively used in a wide variety of fields including financial trading~\cite{markowitz1957solution}, cryptography~\cite{odlyzko1990rise} and resource distribution~\cite{vanderster2009resource}. Recent works utilize deep neural networks for efficient and accurate optimization for solving the knapsack problem~\cite{gu2018pointer, martini2019application} To the best of our knowledge, this work is the first to utilize a Knapsack Problem to improve deep neural networks pruning.

\section{Methodology to Prune Sequential Convolutions}
\label{sec:methodology_seq}


\vspace{-5pt} 
In this section, we present our method for pruning sequential convolutions. This allows us to prune networks such as VGG as well as all the convolutions inside ResNet that are not preceding a skip-connection. 
 Generalization to non-sequential operations such as skip-connections or integration of operations, is presented in Section~\ref{sec:gmcc}.

\vspace{-5pt} 
\subsection{Knapsack Problem and Pruning}
\vspace{-5pt} 
Suppose we have a knapsack with a capacity $C$ and a collection of $n$ items $\mathcal{I}$ where every item $o_i\in\mathcal{I}$ has a weight $f_i$ and a value $v_i$. 
The Knapsack Problem aims to fill the knapsack with maximal value, considering the weight capacity $C$. 
That is
\begin{align}
\max_{\boldsymbol{b}}\  & \sum_{i}v_{i}b_{i} \label{eqn:ksp} \\
\mathrm{s.t} \ & \sum_{i}f_{i}b_{i}\leq C, \ b_{i}\in\{0,1\}\quad 1\leq i \leq n \nonumber 
\end{align}
where the indicator variable $b_i$ equals $1$ if $o_i$ is selected and $0$ otherwise. 

The above formulation is an integer programming problem which is NP-hard. 
If the weights $f_i$ are integers, the problem has an exact solution that can be found with a Dynamic Programming algorithm in a $O\left(n\max_if_i\right)$ time complexity. An approximate solution of the problem can also be found with a greedy approximation algorithm ~\cite{knapsack} in $O(n\log(n))$ time complexity. The method relaxes the original problem by replacing the constraint  $b_i\in\{0,1\}$ with $0\leq b_i \leq 1$ . Then the approximated solution can be derived in a closed form.


We formulate the network pruning task as a approximate Knapsack problem. Given a network $\mathcal{N}$ with convolutional layers $\mathcal{C}_l, 1\!\leq\!l\!\leq\!L$, we seek to prune its output channels with the least impact on the classification accuracy under a target FLOPs budget $C$. Denote by $\mathcal{P}_\mathcal{N}$ the space of pruned versions of $\mathcal{N}$ and by $\operatorname{Acc}$ the accuracy on a validation set $\mathcal{X}$. We formulate the problem as follows:
\begin{align}
\max_{\mathcal{N}_\text{pruned}} \ & \operatorname{Acc} ( \mathcal{N}_\text{pruned},  \mathcal{X}) \label{eqn:prna}\\ 
\mathrm{s.t} \ & \mathcal{N}_\text{pruned} \in \mathcal{P}_\mathcal{N}, \  \operatorname{FLOPs}(\mathcal{N}_\text{pruned})\leq C \nonumber
\end{align}

Optimizing the above problem is not straightforward as the accuracy $\operatorname{Acc}$ is not differentiable.
Therefore, it is common to use an approximated formulation that minimize the cross-entropy loss to replace the $\operatorname{Acc}$.
%
%

Yet, Eq.~\eqref{eqn:prna} remains costly to solve, therefore we next propose an additional approximation. Instead of maximizing the accuracy (minimizing the cross-entropy loss), we minimize the change of the loss due to zeroing-out the pruned network neurons. 
Correspondingly, we adjust the constraint of FLOPs to 
constrain the accumulated FLOPs that are associated with the selected weights. 
The space $\mathcal{P}_\mathcal{N}$ can be represented with a binary indicator vector $\mathbf{b}$ where $b_i\!\in\!\{0,1\}$ indicates if the network's weight $w_i$ is zero or not. We denote by $I(w_i)$ the change of the loss $\mathcal{L}_\text{CE}(x,{\mathcal{N}_\text{pruned}})$ and by $F(w_i)$ the saving of the FLOPs when setting $b_i$ to zero.  
Problem~\eqref{eqn:prna} can be now approximated as:
\begin{align}
\max_{\vb} \ & \sum_i b_i I(w_i)  \label{eqn:kspp} \\
\mathrm{s.t} \ & \sum_i b_i F(w_i) \leq C, \ b_i \in\{0,1\} \ \forall i \nonumber
\end{align}
The above Eq. (\ref{eqn:kspp}) is equivalent to the Knapsack Problem Eq.~\eqref{eqn:ksp}. We will now describe how we compute $I(w_i)$ and $F(w_i)$.

The change of loss $I(w_i)$ can be approximated by the first order Taylor Expansion of the loss function ~\cite{DBLP:conf/iclr/MolchanovTKAK17}. 
Formally, given a function $f:\mathbb{R}^n\rightarrow \mathbb{R}$ and a vector $\vw\in \mathbb{R}^n = \sum_i w_i \ve_i$ where $\ve_i$ is the $i$-th canonical vector of $\mathbb{R}^n$ filled with $0$ everywhere except for the $1$-th coordinate. Denote $\tilde{\vw}^j=\sum_{i\neq j}w_i \ve_i$ a copy of the vector $\vw$ with the $j$-th coordinate replaced by zero. We have
$$
f(\tilde{\vw}^j)=f(\sum_{i\neq j}w_i \ve_i) \approx f(\vw)-w_j\frac{\partial f(\vw)}{\partial w_j}.
$$
Therefore the impact on the loss of zeroing the weight $w_l^o$ of the $o$-th output channel of the $l$-th layer can be approximated by:
\begin{equation}
\label{eqn:Taylor_loss_before}
I(w_l^o)\approx- \mathbb{E}_{\mathbf{x}}\left( {w_l^o}^T\frac{\partial \mathcal{L}(\mathbf{x}, \mathbf{w})}{\partial w_l^o}\right)
\end{equation}
where $\mathbf{x}$ is the input instances (images for example).
The higher this value, the higher the impact of the weight on the total loss. Unfortunately, the above approximation may be too noisy  since the expectation of the gradient is zero at the convergence point of the loss function. In \cite{DBLP:conf/iclr/MolchanovTKAK17}, they show that the variance of the quantity ${z_l^o}={w_l^o}^T\frac{\partial \mathcal{L}(\mathbf{x}, \mathbf{w})}{\partial w_l^o}$ is usually non-zero and correlates with the stability of the local function with respect to $w_l^o$ proposing the following approximation instead:
\begin{equation}
\label{eqn:Taylor_loss_m}
I(w_l^o)\approx \mathbb{E}_{\mathbf{x}}  \left|{w_l^o}^T\frac{\partial \mathcal{L}(\mathbf{x}, \mathbf{w})}{\partial w_l^o}\right|.
\end{equation}

Empirically, we observe that using the below approximation leads to better performances:
\begin{equation}
\label{eqn:Taylor_loss_m2}
I(w_l^o)\approx \mathbb{E}_{\mathbf{x}} \left( \left|{w_l^o}^T\right|\left|\frac{\partial \mathcal{L}(\mathbf{x}, \mathbf{w})}{\partial w_l^o}\right|\right).
\end{equation}

In practice, the expectation in Eq.~\eqref{eqn:Taylor_loss_before} can be approximated by averaging over a validation set. 

Last, we need a formula to calculate the saving of FLOPs $F(w_i)$ after removing the network weight $w_l^o$. Up to now, we focus on the single weight. But in pruning we remove weights in groups. More particularly, we remove a group of weights that are used to compute a channel, such as a filter in a common convolutional layer. Given a convolution with $C_i^l$ input channels of size $H^l\!\times\!W^l$ and $C_o^l$ output channels with kernel size $k^l\!\times\!k^l$ and stride $s^l$, its FLOPs is $C_o^lC_i^lH^lW^l({k^l})^2/({s^l})^2$. 
Zeroing a group of weights related to $w^o_l$ requires removing both an output channel from layer $\mathcal{C}_l$ and an input channel from layer $\mathcal{C}_{l+1}$. Therefore, the saving of FLOPs is given by 
\begin{equation}
\label{eqn:flops}
F(w^o_l)=\frac{C_i^lH^lW^l({k^l})^2}{({s^l})^2} + \frac{C_o^{l+1}H^{l+1}W^{l+1}({k^{l+1}})^2}{({s^{l+1}})^2}.
\end{equation}

Solving the Knapsack Problem~\eqref{eqn:kspp} could be done via dynamic programming.
The complexity of the dynamic programming is $O(nF_\text{max})$, and in our case, $F_\text{max}=\max_i F(w^o_l)$ represents the maximum FLOPs required by a convolutional channel of the network, and can be computed with Eq.~\eqref{eqn:flops}. 
In practice, we can reduce the computational complexity from $O(nF_\text{max})$ to $O\left(\frac{nF_\text{max}}{g}\right)$, 
where $g$ is the Greatest Common Divisor (GCD) of the set $\{F(w^o_l)~ \forall 1\!\leq\!l\!\leq\!L$\}. Dividing both $F(w^o_l)$ and $C$ by $g$ accelerates the convergence time without changing the solution. The total knapsack runtime is negligible in comparison to the network fine-tuning process discussed in Section \ref{sec:IKD}. 
The details of the optimization procedure are described in Algorithm (1) in the supplementary. In addition, we can replace the FLOPS constraint by a running time constraint. In the supplementary material, we present the results of some networks trained with such a method, and show that our formulation allows to get time-pruned networks with the highest accuracy for a given inference time constraint. 

\vspace{-10pt} 
\section{Pruning Non-Sequential Convolutions}
\vspace{-5pt} 
\label{sec:gmcc}
To date, most pruning methods are restricted to sequential connections as non-sequential connections are non trivial to prune.

We next suggest a method that allows pruning of non-sequential connections as part of the proposed knapsack framework. 
The key idea is to group operations that together directly form a channel or a group of channels in a feature map. For example all the convolutions whose outputs are connected through a summation, a multiplication or any inherent constraint like the one in separable convolution.
In this setting, the channels of every group are pruned together, and the pruned network structure is consistent with the unpruned one.

\begin{wrapfigure}{T}{0.5\textwidth}

    \centering
\tikzstyle{conv} = [rectangle, draw, fill=white!20, 
    text width=2.5em, text centered, rounded corners, minimum height=3.5em]
\tikzstyle{in_out} = [text centered, rounded corners, minimum height=1.5em]
\tikzstyle{plus} = [circle, draw, fill=white!20, text centered, rounded corners]

\tikzstyle{mult} = [circle, draw, fill=gray!20, text centered, rounded corners]
\tikzstyle{conv_se} = [rectangle, draw, fill=gray!20, 
    text width=2.5em, text centered, rounded corners, minimum height=3.5em]

\tikzstyle{se} = [circle, draw, fill=gray!20, 
    text width=2.0em, text centered, rounded corners, minimum height=2.0em]
\tikzstyle{line} = [draw, -latex]
\tikzstyle{every node}=[font=\tiny]
\begin{tikzpicture}[node distance = 1.5cm, auto]
    \node [in_out] (input) {input};
    \node [conv, right of=input, node distance=1cm] (pw_e) {Point-wise expansion};
    \node [conv, right of=pw_e] (dw) {Depth-wise convolution};
    \node [se, right of=dw] (se) {Squeeze-and-Excitation};
    \node [conv, right of=se] (pwl) {Point-wise linear projection};
    \node [plus, right of=pwl, node distance = 1cm] (plus) {+};
    \node [in_out, right of=plus, node distance = 1cm] (out) {output};
    
    \node [conv_se, below of=pw_e, node distance = 2cm] (ap) {Average Pooling};
    \node [conv_se, below of=dw, node distance = 2cm] (rd) {Reduction Convolution};
    \node [conv_se, below of=se, node distance = 2cm] (ed) {Expansion Convolution};
    \node [conv_se, below of=pwl, node distance = 2cm] (gate) {Gate};
    \node [mult, below of=plus, node distance = 2cm] (mult) {*};
    
    
    \path [line] (ap) -- (rd);
    \path [line] (rd) -- (ed);
    \path [line] (ed) -- (gate);
    \path [line] (gate) -- (mult);

    \path [line] (input) -- (pw_e);
    \path [line] (pw_e) -- (dw);
    \path [line] (dw) -- (se);
    \path [line] (se) -- (pwl);
    \path [line] (pwl) -- (plus);
    \path [line] (plus) -- (out);
    \path [line] (input.north)  -- ++(0,0.5)  -- ++(0,0) -| (plus);
    
    \path [line] (ap.north)  -- ++(0,0.5)  -- ++(0,0) -| (mult);

\end{tikzpicture}
    \caption{Inverted Residual Block with Squeeze-and-Excitation}
    \label{fig:inverted_res}
\end{wrapfigure}
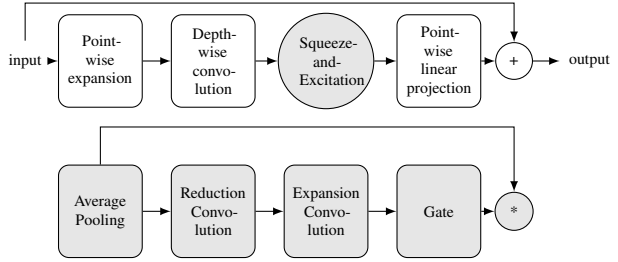

To make this more clear we take as an example a cell called {\it inverted residual} as shown in Figure~\ref{fig:inverted_res} where we neglect activation functions for brevity.
This cell appears in EfficientNet~\cite{EfficientNet}, MNASnet~\cite{Mnas} and MobileNet~\cite{mobilenetv3}.
This cell contains both Squeeze-and-Excitation components~\cite{hu2018senet} and dilated convolutions. 

There are three constraints on the inverted residual block.
First, the output channels of the 'Point-wise linear projection' have to match the input of the current block because of the skip-connection. Second, the output channels of the 'Point-wise expansion' have to match the output channels of the 'Depth-wise convolution' since a Depth-wise convolution has a number of output channels that corresponds the the number of input channels. Lastly, the output channels of the 'Depth-wise convolution' have to match the output channels of the 'Squeeze-and-Excitation Expansion Convolution' because of the skip multiplication.

In order to prune this cell we build three groups of convolutions. 
The first includes the successive 'Point-wise linear projections' of the different blocks. The second includes the 'Point-wise expansions', the 'Depth-wise convolutions' and 'Squeeze-and-Excitation Expansion convolutions' of the same block. 
The third consists of the 'Squeeze-and-Excitation Reduction Convolutions'.
As mentioned above, for each of these three groups we prune their associated channels together.

To the best of our knowledge, we are the first to suggest a pruning method that applies effectively to a non-sequential architecture such as EfficientNet.

\vspace{-5pt} 
\section{Inner Knowledge Distillation and Fine-Tuning}
\label{sec:IKD}
\vspace{-5pt} 
After we get the architecture of the pruned network, we fine-tune its weights. Here we present a method that accelerates the process of fine-tuning by reducing the number of steps. For instance, in TAS~\cite{TAS}, they require 236 GPU hours to search for the pruned version of ResNet-18 using NVIDIA Tesla V100 GPUs. Our method finds the pruned network in less that 0.1 GPU hours and requires 19 GPU hours using the same NVIDIA Tesla V100 GPUs to fine-tune the network. That is 12 times faster.

A common practice in fine-tuning is to incorporate a Knowledge Distillation term \cite{KnowledgeDistillation, tian2020contrastiveRepDistillation} in the loss function. 
This has proven to be very efficient and increases the final accuracy of a student network when using a high accuracy teacher network.

Denote by $\mathcal{N}_{\text{Teacher}}, \mathcal{N}_{\text{Student}}$, the teacher and student networks, and their respective output logits by $\mathcal{F}^t_\text{out},\mathcal{F}^s_\text{out}$. 
Let $\operatorname{SM}(\cdot)$ denote the softmax operator defined by
$
\operatorname{SM}(\vy)_i=\frac{\exp(y_i)}{\sum_j \exp(y_j)}.
$
The KD enforces the output logits distributions of the teacher and student networks to be as similar as possible.  This is achieved by adding  Kullback–Leibler divergence in the loss function as
\begin{equation}
\label{eqn:LKD}
\mathcal{L}_\text{KD}=\sum_{x,i} -\log\left(\operatorname{SM}\left(\mathcal{F}^s_\text{out}(x)\right)_i \right)\operatorname{SM}\left(\mathcal{F}^t_\text{out}(x)\right)_i.
\end{equation}

\vspace{-10pt}We next suggest a further loss term that aims for similarity between $\mathcal{N}_{\text{Teacher}}$ and $\mathcal{N}_{\text{Student}}$, not only between their output logits but also between their internal feature maps.

A distillation at the level of the feature maps between two different networks has been already studied by previous works such as \cite{fitnets, Heo_2019_ICCV}, but the internal feature maps on which the distillation is performed are chosen arbitrarily, since the teacher and student networks have different structures. In the scope of pruning, we do not have this limitation since the teacher and student networks have the same exact structure up to the number of channels in every convolution. As far as we know, we are the first to use a feature maps distillation on pruning method. What allows us to perform such a distillation is the fact that our method is one-shot, meaning that we choose only once the channel to be pruned, unlike other iterative methods such as \cite{CCP, FPGM, TAS} where the choice of the weights to be pruned is constantly updated during the process.

Let $\mathcal{F}^t_l$ be the output feature map at the $l$-th layer of $\mathcal{N}_{\text{Teacher}}$ with $C_l^t$ channels.
Similarly, the output feature map at the $l$-th layer of $\mathcal{N}_{\text{Student}}$ is $\mathcal{F}^s_l$ with $C_l^s$ channels.
In our case, $\mathcal{N}_{\text{Teacher}}$ and $\mathcal{N}_{\text{Student}}$ have the same structure apart from their convolutional channel numbers. 
Hence we could transfer the knowledge inside the network at the level of the convolutional layers. Since the convolution before activation is a linear operator, we require the pruned network to reconstruct the original feature map. We call this the Inner Knowledge Distillation (IKD). Mathematically, we define the IKD loss term as
\begin{equation}
\label{eqn:LIKD}
\mathcal{L}_\text{KD}=\sum_x \left(\sum_l \left\| \mathcal{F}_l^t(x,W_t)-\mb{M}_l\mathcal{F}_l^s(x,W_s)\right\|^2_2 \right)
\end{equation}
where $W_l$ represents the weight matrix at layer $l$ and $\mb{M}_l$ is a $(C^t_l \times C^s_l )$ matrix that aims to reconstruct the features maps $\mathcal{F}^t_l$ from $\mathcal{F}^s_l$, and is added to the list of learnable variables in the fine-tuning process. To avoid degenerate solutions, we add a weight decay regularization term to $\mb{M}_l$, that behaves like a ridge regression regularizer.

The final loss used in the fine-tuning combines the original cross-entropy loss $\mathcal{L}_\text{CE}$, the Knowledge Distillation loss (\ref{eqn:LKD}) and the Inner Knowledge Distillation loss (\ref{eqn:LIKD}):
\begin{equation}
\label{eqn:fL}
\mathcal{L}=\mathcal{L}_\text{CE}+\lambda_{\text{IKD}}\mathcal{L}_\text{IKD}+\lambda_{\text{KD}}\mathcal{L}_\text{KD}
\end{equation}


\vspace{-15pt} 
\section{Experiments}
\vspace{-5pt} 
\label{sec:exp}
In this section, we present empirical results of our pruning method on three different benchmarks: ImageNet \cite{imagenet_cvpr09}, CIFAR-10 and CIFAR-100 \cite{CIFAR}.
To show robustness to the architecture, we experiment with a variety of depths of ResNet~\cite{ResNet} as well as EfficientNet~\cite{EfficientNet}. We further present more experiments on ImageNet since this benchmark is more challenging than CIFAR, and is the standard benchmark for evaluating modern networks. 

The experimental protocol is as follows: 
We first train a full-size baseline network on the selected dataset, next we prune it using our Knapsack formulation and last we apply fine-tuning with IKD {\bf{for 50 epochs only}}, even though most of the other methods fine-tune for more than 100 epochs.


\vspace{-5pt} 
\subsection{ImageNet}
\vspace{-5pt} 
\subparagraph{Comparison to other pruning methods}
To test our method on ImageNet, we used three versions of ResNet~\cite{ResNet}: ResNet-18, ResNet-50, and ResNet-101.

\begin{figure}
\begin{subfigure}[t]{.49\textwidth}
\begin{tikzpicture}[scale=.8]
\centering
\begin{axis}[
    title={Accuracy vs FLOPs on ImageNet},
    xlabel={FLOPs (Billions)},
    ylabel={Top-1 Accuracy (\%)},
    xmin=0, xmax=9,
    ymin=68, ymax=82,
    xtick={1,2,3,4,5,6,7,8,9},
    ytick={68,69,70,71,72,73,74,75,76,77,78,79,80,81,82},
    legend pos=south east,
    ymajorgrids=true,
    xmajorgrids=true,
    grid style=dashed,
    legend style={font=\tiny},
]
\addplot[
    color=red,
    mark=o,
    ]
    coordinates {
    (1.37, 77.50)(1.81, 78.36)(2.48, 79.17)(3.59, 79.74)(5.44, 80.24)(8.06, 80.42)
    };
    \addlegendentry{ResNet-101-Pruned}
 \addplot[
    only marks,
    color=red,
    mark=*]
    coordinates {
    (8.06, 80.42)
    } node (ResNet-101){};
    \addlegendentry{ResNet-101-Baseline}
\addplot[
    color=blue,
    mark=o,
    ]
    coordinates {
    (1.64, 76.81)(2.05, 77.80)(2.46, 78.20)(2.79, 78.27)(4.14, 78.47)
    };
    \addlegendentry{ResNet-50-Pruned}
     \addplot[
    only marks,
    color=blue,
    mark=*]
    coordinates {
    (4.14, 78.47)
    } node (ResNet-50){};
    \addlegendentry{ResNet-50-Baseline}
\addplot[
    color=ForestGreen,
    mark=o,
    ]
    coordinates {
	(1.09, 69.35)(1.17, 69.96)(1.34, 70.61)(1.83, 71.19)
    };
    \addlegendentry{ResNet-18-Pruned}
     \addplot[
    only marks,
    color=ForestGreen,
    mark=*]
    coordinates {
    (1.83, 71.19)
    } node (ResNet-18){};
    \addlegendentry{ResNet-18-Baseline}

\node[color=blue, above of=ResNet-50, font=\tiny, node distance = 0.2cm] {ResNet-50};  
\node[color=red, below of=ResNet-101, font=\tiny, node distance = 0.2cm] {ResNet-101};
\node[color=ForestGreen, below of=ResNet-18, font=\tiny, node distance = 0.2cm, xshift=1em] {ResNet-18};  


\addplot[only marks,color=black,mark=*]coordinates {(3.5,76.2)} node (DenseNet169){};    
\node[color=black, below of=DenseNet169, font=\tiny, node distance = 0.2cm] {DenseNet169};  

\addplot[only marks,color=black,mark=*]coordinates {(4.3,77.42)} node (DenseNet201){};    
\node[color=black, below of=DenseNet201, font=\tiny, node distance = 0.2cm] {DenseNet201};

\addplot[only marks,color=black,mark=*]coordinates {(6,77.9)} node (DenseNet264){};    
\node[color=black, below of=DenseNet264, font=\tiny, node distance = 0.2cm] {DenseNet264};  

\addplot[only marks,color=black,mark=*]coordinates {(5.7,78.8)} node (Inception-v3){};    
\node[color=black, below of=Inception-v3, font=\tiny, node distance = 0.2cm] {Inception-v3};  

\addplot[only marks,color=black,mark=*]coordinates {(1.94,74.8)} node (Inception-v2){};    
\node[color=black, below of=Inception-v2, font=\tiny, node distance = 0.2cm] {Inception-v2};  

\addplot[only marks,color=black,mark=*]coordinates {(8.4,79.0)} node (Xception){};    
\node[color=black, below of=Xception, font=\tiny, node distance = 0.2cm] {Xception};  

\addplot[only marks,color=black,mark=*]coordinates {(4.2,77.8)} node (ResNext50){};    
\node[color=black, left of=ResNext50, font=\tiny, node distance = 0.6cm] {ResNext50};

\end{axis}
\end{tikzpicture}
\caption{Comparison of deep pruned and shallower unpruned networks. Pruning ResNet-101 provides a network with less FLOPs and better accuracy than other networks.}

\label{fig:acc}
\end{subfigure}
\hfill
\begin{subfigure}[t]{.49\textwidth}
\begin{tikzpicture}[scale=.8]
\begin{axis}[
            title={Accuracy vs FLOPs of ResNet-50 on ImageNet},
            axis x line=left,
            axis y line=left,
            xmajorgrids=true,
            ymajorgrids=true,
            grid=both,
            enlarge x limits,
            ymin = 74.25,
            ymax = 78.5,
            xmin = 2.1,
            xmax = 3.0,
            xtick = {2.1,2.2,...,3.0},
            ytick = {74.25,74.5,...,78.5},
            ylabel = Top-1 Accuracy (\%),
            xlabel = FLOPs  (Billions),
            legend pos=north east,
    ]

\draw[very thick, dashed, color=red](1.4,78.45) -- (3.5,78.45);
\draw[very thick, dashed, color=blue] (1.4,76.15) -- (3.5,76.15);
\draw[very thick, dashed, color=ForestGreen] (1.4,77.46) -- (3.5,77.46);

\addplot[only marks,color=gray,mark=triangle*,very thick]coordinates {(2.05, 177.70)};\addlegendentry{Ours}

\addplot[only marks,color=gray,mark=*,very thick]coordinates {(2.38, 74.61)};\addlegendentry{Others}

\addplot[only marks,color=red,mark=triangle*,very thick]coordinates {(2.46,78.2)(2.30, 78.02)(2.05, 77.80)};

\addplot[only marks,color=blue,mark=triangle*,very thick]coordinates {(2.05, 76.21)(2.38, 76.60)};

\addplot[only marks,color=blue,mark=*,very thick]coordinates {(2.38, 74.61)} node (SFP){};

\node[color=blue, below of=SFP, font=\tiny, node distance = 0.2cm] {SFP};  

\addplot[only marks,color=blue,mark=*,very thick]coordinates {(2.25, 74.50)} node (Taylor){};
\node[color=blue, below of=Taylor, font=\tiny, node distance = 0.2cm] {Taylor};

\addplot[only marks,color=blue,mark=*,very thick]coordinates {(3.00, 76.00)} node (AutoSlim){};
\node[color=blue, below of=AutoSlim, font=\tiny, node distance = 0.2cm] {AutoSlim};

\addplot[only marks,color=blue,mark=*,very thick]coordinates {(2.36, 75.50)} node (FPGM){};
\node[color=blue, above of=FPGM, font=\tiny, node distance = 0.2cm] {FPGM};

\addplot[only marks,color=red,mark=*,very thick]coordinates {(2.36, 76.4)} node (FPGM2){};
\node[color=red, above  right of=FPGM2, font=\tiny, node distance = 0.3cm] {FPGM};


\addplot[only marks,color=blue,mark=*,very thick]coordinates {(2.66, 75.48)} node (TaylorFoBn){};
\node[color=blue, below of=TaylorFoBn, font=\tiny, node distance = 0.2cm] {Taylor-FO-BN};

\addplot[only marks,color=blue,mark=*,very thick]coordinates {(2.30, 74.90)} node (Slimable){};
\node[color=blue, below of=Slimable, font=\tiny, node distance = 0.2cm] {Slimable};

\addplot[only marks,color=blue,mark=*,very thick]coordinates {(2.13, 75.50)} node (CCP){};
\node[color=blue, above of=CCP, font=\tiny, node distance = 0.2cm] {CCP};

\addplot[only marks,color=blue,mark=*,very thick]coordinates {(2.58, 75.53)} node (AOFPC1){};
\node[color=blue, above of=AOFPC1, font=\tiny, node distance = 0.2cm] {AOFP-C1};

\addplot[only marks,color=ForestGreen,mark=*,very thick]coordinates {(2.31, 76.20)} node (TAS){};
\node[color=ForestGreen, above of=TAS, font=\tiny, node distance = 0.3cm] {TAS};

\addplot[only marks,color=ForestGreen,mark=triangle*,very thick]coordinates {(2.31, 77.50)};

\addplot[only marks,color=ForestGreen,mark=triangle*,very thick]coordinates {(2.03, 77.06)};

\end{axis}
\end{tikzpicture}
\caption{Impact of the baseline. Every color is assigned to a different baseline. Red, blue and green entries are respectively from our, PyTorch and TAS baseline (dotted lines). }
\label{fig:acc_pruning}
\end{subfigure}
\caption{Top-1  accuracy v.s. FLOPs for pruned ResNets on ImageNet.}
\end{figure}
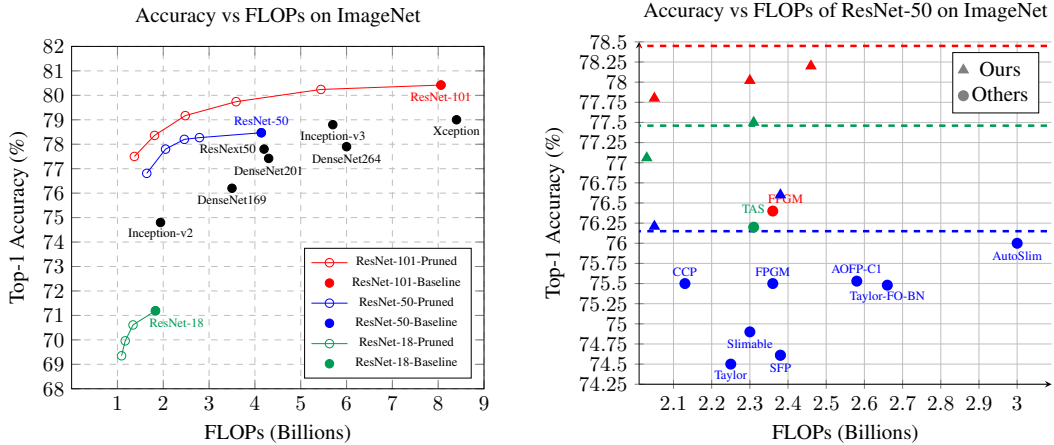

Table~\ref{table:ImageNet-SOTA} and Figure~\ref{fig:acc_pruning} compare our results for different pruning ratios with previous works.
It can be seen that our results are consistently better than the state-of-the-art. 

\vspace{-5pt} 
\subparagraph{Comparison to common classification networks}
To further evaluate the benefits of our pruning approach we present in Figure~\ref{fig:acc} a comparison of the performance of our pruned networks with popular architectures: Inception~\cite{InceptionV2}, DenseNet~\cite{DensNet}, ResNext~\cite{ResNext} and Xception~\cite{Xception}.
We compare both Top-1 accuracy and computational cost (FLOPs). 
It can be seen that our pruned networks consistently provide higher accuracy than other networks, for a given number of FLOPs. 
\vspace{-5pt} 
\subparagraph{Ablation study}
\begin{wraptable}{T}{0.5\textwidth}
  \caption{
  Ablation study of pruned ResNet-50 on ImageNet.
  }
  \centering
  \setlength{\tabcolsep}{2pt}
  \resizebox{\linewidth}{!}{
  \begin{tabular}{c c c c c c}
    \toprule
  Baseline & IKD & Prune Acc   &  Acc Drop & FLOPs & \makecell{Prune Ratio $\downarrow$} \\
    \midrule
    High & \cmark  &  78.20\%     &  0.27\%    &  \multirow{2}{*}{2.46E9} &   \multirow{2}{*}{40.64\%} \\
    High & \xmark     &  77.12\%     &  1.35\%    &   &    \\
      PyTorch& \cmark &  76.60\%     &  -0.46\%  &  \multirow{2}{*}{2.38E9} &   \multirow{2}{*}{42.56\%} \\
      PyTorch & \xmark
      &  76.17\%     &  -0.03\%  &  &    \\
      
    \hdashline[5pt/5pt]
    High & \cmark     &  77.82\%     &  0.65\%    &  \multirow{2}{*}{2.05E9} &   \multirow{2}{*}{50.50\%} \\
    High & \xmark   &  76.70\%     &  1.77\%    &  &   \\
   PyTorch & \cmark  &  76.21\%     &  -0.07\%  &  \multirow{2}{*}{2.03E9} &   \multirow{2}{*}{50.80\%} \\
   PyTorch & \xmark  &  75.94\%     &  0.21\%  &  &  \\

    \bottomrule
  \end{tabular}
  }
  \label{table:ImageNet-ablation}

\end{wraptable}

Next, we present an ablation study, to assess the contribution of the various components of our approach.
We took ResNet-50 as backbone and experimented with two variants: (i) With and without IKD, and (ii) our baseline training vs. PyTorch baseline. Results are presented in Table~\ref{table:ImageNet-ablation}. For a fair comparison with regard to the impact of our baseline, we take the original implementation of FPGM~\cite{FPGM} and prune our own baseline ResNet-50 instead of the original PyTorch one. Next, we prune ResNet-50 using the same baseline of $77.46\%$ top-1 accuracy as TAS\cite{TAS}. In both cases, we can see that our method provides better results, no matter the baseline we start from as can be seen in Figure~\ref{fig:acc_pruning}.

\textit{IKD:\ \ } 
When using IKD, we have more than 1\% improvement than when not using KID, both for pruning 50\% of ResNet-50 and pruning 41\%.
As could be expected, when using as baseline the low-accuracy network provided with PyTorch, the performance improvement by the IKD step is smaller, going from  $76.17\%$ without IKD to $76.60\%$ with IKD.

\textit{Baseline:\ \ }
To measure the impact of the baseline on our method, we choose to prune ResNet-50 with the official PyTorch \cite{PyTorch} pre-trained weights that leads to $76.15\%$ Top-1 accuracy on ImageNet.
This is the common evaluation scheme adopted by most works. 
Comparing our results in Table~\ref{table:ImageNet-ablation} with those of previous work in Table~\ref{table:ImageNet-SOTA} shows that our method still provides the highest accuracy.

\textit{Knapsack:\ \ } To assess the contribution of the Knapsack formulation, we have pruned 42.6\% of ResNet-50 on ImageNet on the official Pytorch baseline using Molchanov's criterion only \cite{molchanov2019taylor}, without the Knapsack formulation and have obtained 75.26\% accuracy, while the addition of the Knapsack formulation (without IKD) led to 76.17\% accuracy, an improvement 0.91\%. This result stands to demonstrate the importance of the Knapsack formulation, and that our results are not due to the fact that we use the Taylor Expansion criterion. 
\vspace{-10pt}\subsection{Pruning the Non-Sequential EfficientNet}
\vspace{-5pt} 
\begin{wraptable}{r}{0.5\textwidth}
\vspace{-25pt}
 \caption{Comparison of pruned and original versions of EfficientNet. Inference speed (images/second) is measured on GPU NVIDIA P100. Similar to our observation on ResNets in Fig.~\ref{fig:acc}, our pruning method consistently provides networks with superior accuracy than other networks with same FLOPS.}
  \centering
  \setlength{\tabcolsep}{1pt}
  \resizebox{\linewidth}{!}{
  \begin{tabular*}{0.6\textwidth}{l @{\extracolsep{\fill}} c c c}
    \toprule
    Network & Acc & FLOPs & Speed (Im/s) \\
    \midrule
    MobileNetV3 Large & 75.2\% & \multirow{2}{*}{0.21E9} & 1730 \\
    EfficientNet B0 Pruned & \textbf{75.5\%} &  & \textbf{2133} \\
    \midrule
    EfficientNet B0 & 77.3\% & \multirow{2}{*}{0.39E9} & 1230 \\
    EfficientNet B1 Pruned & \textbf{78.3\%} &  & \textbf{1355} \\
    \midrule
    EfficientNet B1 & 79.2\% & \multirow{2}{*}{0.7E9} & 784 \\
    EfficientNet B2 Pruned & \textbf{79.9\%} &  & \textbf{882} \\
    \midrule
    EfficientNet B2 & 80.3\% & \multirow{2}{*}{1.0E9} & 595 \\
    EfficientNet B3 Pruned & \textbf{80.8\%} & & \textbf{683} \\
    \midrule
    EfficientNet B3 & 81.7\% & \multirow{2}{*}{1.8E9} & 350 \\
    EfficientNet B4 Pruned & \textbf{81.9\%} & & \textbf{385} \\
    \bottomrule
  \end{tabular*}
  }
  \label{table:EfficientNet}
\end{wraptable}

As described in Section~\ref{sec:gmcc}, our approach can be applied also to prune architectures with non-sequential convolutions and skip-connections such as 
EfficientNet~\cite{EfficientNet}.
To the best of our knowledge, this is the first attempt to prune these types of networks. 

We experimented with 4 variants, comparing pruned EfficientNet B$\{n\}$ with EfficientNet B$\{n-1\}$, where $n\in{\{1,2,3,4\}}$.
For a fair comparison with the unpruned baselines, we followed the published EfficientNet training protocol without IKD. Results are presented in Table~\ref{table:EfficientNet}.
It can be observed that the pruned networks achieve higher accuracy than the baselines with the same number of FLOPs.
An interesting observation is that despite having the same theoretical computational complexity, the pruned networks run faster than the unpruned ones. 
Furthermore, our pruned version of EfficientNet B0 led to a network with the same amount of FLOPs as MobileNetV3-large~\cite{mobilenetv3} and a better accuracy.
\vspace{-10pt}
\subsection{CIFAR}
\vspace{-5pt}
For the CIFARs datasets, we train ResNet-56 on CIFAR-10 and CIFAR-100 according to the same protocol used for ImageNet while changing the number of epochs to 300. 
Our top-1 accuracy baseline is $94.2\%$ for CIFAR-10 and $73.55\%$ for CIFAR-100. Results and comparisons to other works can be seen on the left of Table \ref{table:CIFAR-SOTA}.

\begin{table}[t]
  \caption{
  Comparison of different pruning algorithms for different ResNet backbones on ImageNet.
  }
  \centering
  \setlength{\tabcolsep}{3.5pt}
  \begin{tabular}{c c c c c c c c}
    \toprule
  \multirow{2}{*}{Model}  & \multirow{2}{*}{Method} & \multicolumn{2}{c}{Top-1} & \multicolumn{2}{c}{Top-5} & \multirow{2}{*}{FLOPs} & \multirow{2}{*}{\makecell{Prune\\Ratio}} \\
                          &                         & Prune Acc    &  Acc Drop  & Prune Acc   &    Acc Drop &       &             \\
    \midrule
\multirow{6}{*}{ResNet-18}&LCCL~\cite{LCCL} &  66.33\%     &  3.65\%    &  86.94\%     &   2.29\%   &  1.19E9 &   34.6\%   \\
                          & SFP~\cite{SFP}   &  67.10\%     &  3.18\%    &  87.78\%     &   1.85\%   &  1.06E9 &   41.8\%   \\
                          &FPGM~\cite{FPGM}&  68.41\%     &  1.87\%    &  88.48\%     &   1.15\%   &  1.06E9 &   41.8\%   \\
                          & TAS~\cite{TAS}                 &  69.15\%     &  1.50\%    &  89.19\%     &   0.68\%   &  1.21E9 &   33.3\% \\\cmidrule[0.5pt](lr){2-8} 
                         & Ours                 &  {69.96\%}     &  {1.23\%}    &  {89.60\%}     &   {0.59\%}   &  1.17E9 &   35.77\% \\
                         & Ours                 &  {69.35\%}     &  1.84\%    &  {89.23\%}     &   0.96\%   &  1.09E9 &   40.01\% \\
     \midrule
\multirow{14}{*}{ResNet-50}& SFP~\cite{SFP}   &  74.61\%     &  1.54\%    &  92.06\%     &   0.81\%   &  2.38E9 &   41.8\%   \\
                          & CP~\cite{CP} &    -         &    -       &  90.80\%     &   1.40\%   &  2.04E9 &   50.0\%   \\
                          & Taylor~\cite{Taylor}&  74.50\%     &  1.68\%    &   -   &    - &  2.25E9 &   44.9\%   \\
                          & AutoSlim~\cite{Autoslim} &  76.00\%     &    -       &    -         &     -      &  3.00E9 &   26.6\%   \\
                          & FPGM~\cite{FPGM}& 75.50\%     &  0.65\%    &  92.63\%     &   0.21\%   &  2.36E9 &   42.2\%   \\
                          &SSS~\cite{SSS2018} & 71.82\%     & 4.30\%    &  90.79\%     &   2.07\%  &  2.33E9 &   43.4\%   \\
                          & Taylor-FO-BN~\cite{molchanov2019taylor}& 75.48\%     &  0.70\%     & -     &   -   &  2.66E9 &   35.5\%   \\
                          & Slimable~\cite{yu2018slimmable}& 74.90\%     &  1.20\%     & -     &   -   &  2.30E9 &   44.0\%   \\
                          & CCP~\cite{CCP} & 75.50\%     &  0.65\%    &  92.62\%     &   0.25\%   &  2.13E9 &   48.8\% \\
                          & AOFP-C1~\cite{AOFPC1} & 75.53\%     &  -0.29\%    &  92.69\%     &   -0.13\%   &  2.58E9 &   32.88\% \\
                          & TAS~\cite{TAS}    & 76.20\%     &  1.26\%    &  93.07\%     &   0.48\%   &  2.31E9 &   43.5\%   \\\cmidrule[0.5pt](lr){2-8}
                          & Ours  & {78.20\%}   &  {0.27\%}    &  {93.98\%}     &   {-0.10\%}   &  {2.46E9} &   {40.64\%} \\
                          & Ours  & {78.02\%}   &  {0.45\%}    &  {93.88\%}     &   {0.00\%}   &  {2.30E9} &   {44.47\%} \\
                          & Ours  & {77.80\%}   &  {0.67\%}    &  {93.78\%}     &   {0.10\%}   &  {2.05E9} &   {50.21\%} \\

\midrule
\multirow{7}{*}{ResNet-101}

			 & Taylor-FO-BN~\cite{molchanov2019taylor}& 75.38\%     &  -     & -     &   -   &  2.47E9 &   69.3\%   \\
			 & FPGM~\cite{FPGM}& 77.32\%    & 0.05\%    &  93.56\%     &  0.00\%    &  4.51E9 & 42.2\% \\
			 & RSNLIA~\cite{ye2018rethinking}& 75.27\% &  2.10\%    &  -           &     -      &  4.13E9 & 47.0\% \\  
			 & AOFP-D2~\cite{AOFPC1}& 76.40\% &  0.23\%    &  93.07\%        &     0.22\%      &  3.77E9 & 50.19\% \\  
			 \cmidrule[0.5pt](lr){2-8}
			 &Ours                &  {79.17\%}     & 1.25\%   & {94.54\%} & 0.63\%& {2.48E9}&   {69.21\%}            \\
                          &Ours                &  {78.36\%}     & 2.06\%   & {94.27\%} & 0.90\%& {1.81E9}&   {77.50\%}            \\
                          &Ours                &  {77.56\%}     & 2.86\%   & {93.68\%} & 1.49\%& {1.37E9}&   {83.00\%}            \\

    \bottomrule
  \end{tabular}
  \label{table:ImageNet-SOTA}
\end{table}

\begin{table}[t]
  \caption{
  Comparison of different pruning algorithms for ResNet-56 on CIFAR.
  }
  \centering
  \setlength{\tabcolsep}{4.5pt}
  \begin{tabular}{c c c c c c c}
    \toprule
 \multirow{1}{*}{Method} &       \multicolumn{3}{c}{CIFAR-10}       & \multicolumn{3}{c}{CIFAR-100}            \\
                                                   & Prune Acc  & Acc Drop  &   FLOPs          & Prune Acc   & Acc Drop  &  FLOPs           \\
    \midrule
                          PFEC~\cite{PFEC}&  93.06\%   & -0.02\%   & 9.09E7 (27.6\%) &   $-$      &   $-$     & $-$             \\
                          LCCL~\cite{LCCL} &  92.81\%   & 1.54\%    & 7.81E7 (37.9\%) &  68.37\%   &   2.96\%  &  7.63E7 (39.3\%) \\
                           AMC~\cite{AMC}    &  91.90\%   & 0.90\%    & 6.29E7 (50.0\%) &    $-$     &   $-$     &  $-$            \\
                           SFP~\cite{SFP}   &  93.35\%   & 0.56\%    & 5.94E7 (52.6\%) &  68.79\%   &   2.61\%  &  5.94E7 (52.6\%) \\
                          FPGM~\cite{FPGM}&  93.49\%   & 0.42\%    & 5.94E7 (52.6\%) &  69.66\%   &   1.75\%  &  5.94E7 (52.6\%) \\
                           CCP~\cite{CCP}  &  93.69\%   & -0.19\%    & 6.61E7 (47.0\%) &  -   &   -  &  -  \\
                           TAS~\cite{TAS}                 &  93.69\%   & 0.77\%    & 5.95E7 (52.7\%) &  72.25\%   &   0.93\%  &  6.12E7 (51.3\%)  \\
                           \midrule
                          Ours &  {93.83\%}   & 0.69\%    & {5.79E7 (53.8\%)} &  {72.62\%}   &   {0.93\%}  &  6.25E7 (50.2\%) \\
    \bottomrule
  \end{tabular}
  \label{table:CIFAR-SOTA}
\end{table}
\section{Conclusion}
\vspace{-10pt}
In this paper we have presented a new formulation and method for the pruning task, which enables us to simultaneously optimize over both accuracy and FLOPs measures, as well as distill knowledge from the unpruned network. This method has provided state-of-the-art empirical results on ImageNet and CIFAR datasets, which demonstrate the effectiveness of our proposed solution. 
We have observed that pruning a heavy deep network with our method can provide better accuracy than a shallower one with the same computational complexity (whether the latter was designed with a Network Architecture Search method or manually). 
These findings may suggest that the Network Architecture Search task should focus on finding inflated over-parametrized networks, while leaving the designing of efficient networks for the pruning and knowledge distillation methods.

\ifnum \arxiv>0

\section{Supplementary Material}

\subsection{Pruning of ECA-ResNet-D}
In this section, we present results that we obtained by pruning a neural network superior to ResNet50. We choose to prune ECA-ResNet-D. This network has a backbone of ResNet, but we add two modifications. First, we change the stem cell as presented in \cite{bag_of_tricks}. In addition, we add ECA modules, as suggested in \cite{ecaresnet}.
The obtained architecture performs better than classical ResNet, and thus, pruning this network is more interesting.
We have pruned different versions of ECA-ResNet-D with our method using two criteria: Flops based pruning, as described in the paper, and Inference-time based pruning. In this setting, we measure the inference time of every convolutional layer of the network, and apply our knapsack method where, instead of imposing a constraint on the total FLOPS of the pruned network, we impose a constraint on the final inference time.
Most of the pruning method focus on reducing the total FLOPS of the pruned networks, but this measure does not often reflects the real inference time on a GPU, and networks with few flops such as EfficientNet \cite{EfficientNet} runs with a lower throughput than heavier architecture such as ResNet. 
In the below table, we present the results of pruning several versions of ECA-ResNet-D, with both FLOPS based pruning and time-based pruning.
\begin{table}[H]
  \caption{
  Performance of FLOPS based and Time based pruning of ECA-ResNet on Imagenet Dataset
  }
  \centering
  \begin{tabular}{c c c c c}
    \toprule
 Network
&
Accuracy 
&
Speed (Im/sec) on V100
&
Speed (Im/sec) on P100
&
Flops (Gigas) \\
\midrule
\multicolumn{5}{c}{Unpruned Architectures}\\
    \midrule
Baseline, Resnet50D
&
79.30\%
&
2728
&
791
&
4.35\\
\midrule

ECA Resnet-50D
&
80.61\%
&
2400
&
718
&
4.35\\
\midrule

ECA Resnet-101D
&
82.19\%
&
1476
&
444
&
8.07\\
\midrule

\multicolumn{5}{c}{Flops based pruned}\\
\midrule

ECA Resnet-50D 41\%
&
80.10\%
&
2434
&
924
&
2.56\\
\midrule

ECA Resnet-101D 55\%
&
81.24\%
&
1651
&
642
&
3.61\\
\midrule

ECA Resnet-101D 68\%
&
80.69\%
&
1735
&
717
&
2.58\\
\midrule

\multicolumn{5}{c}{Time based pruned}\\
 \midrule
ECA Resnet-50D 43\%
&
79.71\%
&
3587
&
1200
&
2.53\\
\midrule

ECA Resnet-50D 39\%
&
79.89\%
&
3145
&
1121
&
2.66\\
\midrule

ECA Resnet-50D 21\%
&
80.34\%
&
2653
&
906
&
3.45\\
\midrule

ECA Resnet-101D 57\%
&
80.86\%
&
2791
&
1010
&
3.47\\

    \bottomrule
  \end{tabular}
  \label{table:ECA}
\end{table}

We can see how time-based pruning using our knapsack method provided extremely fast networks. For example, time-pruning of 57\% of ECA-ResNet-101D provided a network with 80.86\% accuracy while inferring at 1010 images per seconds on a P100 GPU. To the best of our knowledge, this is the fastest network of a NVIDIA P100 GPU to get an accuracy above 80\% on ImageNet. The above networks and their checkpoints have been integrated on the famous Ross Wightman repository \cite{rwightman}, and are available to the public.

\subsection{Knapsack Pruning Algorithm}
\begin{algorithm}[H]
\caption{Knapsack Pruning}
\label{alg:pruning}
\begin{algorithmic}
\INPUT $C,w_i~~\forall i$
\FORALL{$0\leq i\leq n$}
    \STATE Compute $I_i\leftarrow I(w_i)$ 
    \STATE Compute $F_i\leftarrow F(w_i)$ 
\ENDFOR
\STATE Compute $G\leftarrow \operatorname{GCD}(F_i)$
\FORALL{$i$}
    \STATE $F_i\leftarrow F_i/ G$
\ENDFOR
\STATE $C\leftarrow C/ G$
\STATE Initialize $T \leftarrow$ $0$-float array of size $2\times C$
\STATE Initialize $K\leftarrow$ \textbf{False}-binary array of size $n\times C$
\FORALL{$i$}
    \STATE $I_{\text{curr}}\leftarrow I_i, F_{\text{curr}}\leftarrow F_i$
    \STATE $i_\text{prev}\leftarrow (i-1)\%2, i_\text{curr}\leftarrow i\%2$
    \FORALL{$0\leq f\leq C$}
        \IF{$f\geq F_i$}
            \STATE $v_1\leftarrow I_i+T[i_\text{prev}][f - F_i]$
        \ELSE
            \STATE $v_1\leftarrow 0$
        \ENDIF
        \STATE $v_2\leftarrow T[i_\text{prev}][f]$
        \IF{$F_i\leq f$ and $v_2\leq v_1$}
            \STATE $T[i_\text{curr}][f]\leftarrow v_1$
            \STATE $K[i][f]\leftarrow $ \textbf{True}
        \ELSE
            \STATE $T[i_\text{curr}][f]\leftarrow v_2$
        \ENDIF
    \ENDFOR
\ENDFOR
\STATE $P\leftarrow []$
\FOR{$i$ from $n$ to $0$ decreasing by $1$}
    \IF{$K[i][F]$ is \textbf{True}}
        \STATE $P\leftarrow [P,i]$
        \STATE $K\leftarrow K-F_{i-1}$
    \ENDIF
\ENDFOR
\OUTPUT P
\end{algorithmic}
\end{algorithm}
\newpage
\subsection{Pruning ratio and pruned output channels for ResNet 50 }

\begin{figure}[htbp]
\centering
\begin{subfigure}{0.7\textwidth}
\centering
\begin{tikzpicture}
\begin{axis}[
	ylabel=Number of output channels,
	enlargelimits=0.05,
	legend style={at={(0.5,-0.15)},
		anchor=north,legend columns=-1},
	ybar stacked,
	bar width=3pt,
	symbolic x coords={
	layer1.0.0, layer1.0.1,
    layer1.1.0, layer1.1.1,
    layer1.2.0, layer1.2.1,
    layer2.0.0, layer2.0.1,
    layer2.1.0, layer2.1.1,
    layer2.2.0, layer2.2.1,
    layer2.3.0, layer2.3.1,
    layer3.0.0, layer3.0.1,
    layer3.1.0, layer3.1.1,
    layer3.2.0, layer3.2.1,
    layer3.3.0, layer3.3.1,
    layer3.4.0, layer3.4.1,
    layer3.5.0, layer3.5.1,
    layer4.0.0, layer4.0.1,
    layer4.1.0, layer4.1.1,
    layer4.2.0, layer4.2.1
	},
	x tick label style={rotate=45,anchor=east, font=\fontsize{4}{6}\selectfont, text width=0.5cm},
	xtick={
	layer1.0.0, layer1.0.1,
    layer1.1.0, layer1.1.1,
    layer1.2.0, layer1.2.1,
    layer2.0.0, layer2.0.1,
    layer2.1.0, layer2.1.1,
    layer2.2.0, layer2.2.1,
    layer2.3.0, layer2.3.1,
    layer3.0.0, layer3.0.1,
    layer3.1.0, layer3.1.1,
    layer3.2.0, layer3.2.1,
    layer3.3.0, layer3.3.1,
    layer3.4.0, layer3.4.1,
    layer3.5.0, layer3.5.1,
    layer4.0.0, layer4.0.1,
    layer4.1.0, layer4.1.1,
    layer4.2.0, layer4.2.1
	},
]
\addplot 
	coordinates {
	(layer1.0.0,45) (layer1.0.1,26)
    (layer1.1.0,11) (layer1.1.1,11)
    (layer1.2.0,29) (layer1.2.1,31)
    (layer2.0.0,61) (layer2.0.1,56)
    (layer2.1.0,38) (layer2.1.1,44)
    (layer2.2.0,27) (layer2.2.1,37)
    (layer2.3.0,25) (layer2.3.1,43)
    (layer3.0.0,172) (layer3.0.1,144)
    (layer3.1.0,136) (layer3.1.1,131)
    (layer3.2.0,146) (layer3.2.1,118)
    (layer3.3.0,79) (layer3.3.1,100)
    (layer3.4.0,157) (layer3.4.1,104)
    (layer3.5.0,167) (layer3.5.1,131)
    (layer4.0.0,452) (layer4.0.1,447)
    (layer4.1.0,512) (layer4.1.1,460)
    (layer4.2.0,506) (layer4.2.1,492)
	};
	
\addplot 
	coordinates {
	(layer1.0.0,19) (layer1.0.1,38)
    (layer1.1.0,53) (layer1.1.1,53)
    (layer1.2.0,35) (layer1.2.1,33)
    (layer2.0.0,67) (layer2.0.1,72)
    (layer2.1.0,90) (layer2.1.1,84)
    (layer2.2.0,101) (layer2.2.1,91)
    (layer2.3.0,103) (layer2.3.1,85)
    (layer3.0.0,84) (layer3.0.1,112)
    (layer3.1.0,120) (layer3.1.1,125)
    (layer3.2.0,110) (layer3.2.1,138)
    (layer3.3.0,177) (layer3.3.1,156)
    (layer3.4.0,99) (layer3.4.1,152)
    (layer3.5.0,89) (layer3.5.1,125)
    (layer4.0.0,60) (layer4.0.1,65)
    (layer4.1.0,0) (layer4.1.1,52)
    (layer4.2.0,6) (layer4.2.1,20)
    };
\legend{Pruned,Unpruned}
\end{axis}
\end{tikzpicture}
\caption{Number of output channels for ResNet 50}
\label{fig:number_of_channels}
\end{subfigure}
\begin{subfigure}{0.7\textwidth}
\centering
\begin{tikzpicture}
\begin{axis}[
	ylabel=Pruning ratio (\%),
	enlargelimits=0.05,
	legend style={at={(0.5,-0.15)},
		anchor=north,legend columns=-1},
	ybar=1pt,
	bar width=3pt,
	symbolic x coords={
	layer1.0.0, layer1.0.1,
    layer1.1.0, layer1.1.1,
    layer1.2.0, layer1.2.1,
    layer2.0.0, layer2.0.1,
    layer2.1.0, layer2.1.1,
    layer2.2.0, layer2.2.1,
    layer2.3.0, layer2.3.1,
    layer3.0.0, layer3.0.1,
    layer3.1.0, layer3.1.1,
    layer3.2.0, layer3.2.1,
    layer3.3.0, layer3.3.1,
    layer3.4.0, layer3.4.1,
    layer3.5.0, layer3.5.1,
    layer4.0.0, layer4.0.1,
    layer4.1.0, layer4.1.1,
    layer4.2.0, layer4.2.1
	},
	x tick label style={rotate=45,anchor=east, font=\fontsize{4}{6}\selectfont, text width=0.5cm},
	xtick={
	layer1.0.0, layer1.0.1,
    layer1.1.0, layer1.1.1,
    layer1.2.0, layer1.2.1,
    layer2.0.0, layer2.0.1,
    layer2.1.0, layer2.1.1,
    layer2.2.0, layer2.2.1,
    layer2.3.0, layer2.3.1,
    layer3.0.0, layer3.0.1,
    layer3.1.0, layer3.1.1,
    layer3.2.0, layer3.2.1,
    layer3.3.0, layer3.3.1,
    layer3.4.0, layer3.4.1,
    layer3.5.0, layer3.5.1,
    layer4.0.0, layer4.0.1,
    layer4.1.0, layer4.1.1,
    layer4.2.0, layer4.2.1
	},
]
\addplot 
	coordinates {
	(layer1.0.0,29.6875) (layer1.0.1,59.375)
    (layer1.1.0,82.8125) (layer1.1.1,82.8125)
    (layer1.2.0,54.6875) (layer1.2.1,51.5625)
    (layer2.0.0,52.34375) (layer2.0.1,56.25)
    (layer2.1.0,70.3125) (layer2.1.1,65.625)
    (layer2.2.0,78.90625) (layer2.2.1,71.09375)
    (layer2.3.0,80.46875) (layer2.3.1,66.40625)
    (layer3.0.0,32.8125) (layer3.0.1,43.75)
    (layer3.1.0,46.875) (layer3.1.1,48.828125)
    (layer3.2.0,42.96875) (layer3.2.1,53.90625)
    (layer3.3.0,69.140625) (layer3.3.1,60.9375)
    (layer3.4.0,38.671875) (layer3.4.1,59.375)
    (layer3.5.0,34.765625) (layer3.5.1,48.828125)
    (layer4.0.0,11.71875) (layer4.0.1,12.6953125)
    (layer4.1.0,0.0) (layer4.1.1,10.15625)
    (layer4.2.0,1.171875) (layer4.2.1,3.90625)
	};
	
\legend{Pruning ratio}
\end{axis}
\end{tikzpicture}
\caption{Pruning ratio of output channels for ResNet 50}
\label{fig:ratio_of_channels}
\end{subfigure}
\caption{Pruning ratio and pruned output channels for ResNet 50}
\label{fig:subfigureExample}
\end{figure}
\fi

\bibliographystyle{abbrv}
\bibliography{pruning_paper}
\end{document}

%% file: prel.tex
\newcommand{\mb}[1]{{\boldsymbol{#1}}}
\newcommand{\vw}{\mb{w}}
\newcommand{\vb}{\mb{b}}

\newcommand{\ve}{\mb{e}}
\newcommand{\vy}{\mb{y}}